%% file: CVPR2021_arXiv.tex
\documentclass[10pt,twocolumn,letterpaper]{article}

\usepackage{cvpr}
\usepackage{times}
\usepackage{epsfig}
\usepackage{graphicx}
\usepackage{amsmath}
\usepackage{amssymb}
\usepackage{subfigure}
\usepackage{stfloats}
\usepackage{booktabs}
\usepackage{diagbox}
\usepackage{multirow}
\usepackage{graphicx}
\usepackage{color}
\usepackage{makecell}
\usepackage{marvosym}

\usepackage[pagebackref=true,breaklinks=true,letterpaper=true,colorlinks,bookmarks=false]{hyperref}

\cvprfinalcopy 

\def\cvprPaperID{****} 

\ifcvprfinal\fi
\begin{document}

\title{A Fourier-based Framework for Domain Generalization}

\author{
Qinwei Xu\textsuperscript{1} \quad Ruipeng Zhang\textsuperscript{1} \quad Ya Zhang\textsuperscript{1,2 \Letter} \quad Yanfeng Wang\textsuperscript{1,2} \quad Qi Tian\textsuperscript{3} \vspace{0.2cm}
\\ 
\textsuperscript{1} Cooperative Medianet Innovation Center, Shanghai Jiao Tong University \\
\textsuperscript{2} Shanghai AI Laboratory \quad \quad \quad \textsuperscript{3} Huawei Cloud \& AI
\\
{\tt\small $\{$qinweixu, zhangrp, ya\_zhang, wangyanfeng$\}$@sjtu.edu.cn,  tian.qi1@huawei.com}
}

\maketitle
\thispagestyle{empty}

\begin{abstract}
Modern deep neural networks suffer from performance degradation when evaluated on testing data under different distributions from training data. Domain generalization aims at tackling this problem by learning transferable knowledge from multiple source domains in order to generalize to unseen target domains. This paper introduces a novel Fourier-based perspective for domain generalization. The main assumption is that the Fourier phase information contains high-level semantics and is not easily affected by domain shifts. To force the model to capture phase information, we develop a novel Fourier-based data augmentation strategy called amplitude mix which linearly interpolates between the amplitude spectrums of two images. A dual-formed consistency loss called co-teacher regularization is further introduced between the predictions induced from original and augmented images. Extensive experiments on three benchmarks have demonstrated that the proposed method is able to achieve state-of-the-arts performance for domain generalization.    
\end{abstract}

\section{Introduction}

Over the past few years, deep learning have made tremendous progress on various tasks. Under the assumption that training and testing data share the same distribution, deep neural networks (DNNs) have shown great promise for a wide spectrum of applications ~\cite{lecun2015deep, goodfellow2016deep, he2016deep}. However, DNNs have demonstrated quite poor generalizability for out-of-distribution data. Such performance degeneration caused by distributional shift (a.k.a. domain shift) impairs the applications of DNNs, as in reality training and testing data often come from different distributions. 

In order to address the problem of domain shift, domain adaptation (DA) bridges the gap between source domain(s) and a specific target domain with the help of some labelled or unlabeled target data. However, DA methods fail to generalize to unknown target domains that have not been seen during training. Collecting data from every possible target domain and training DA models with every source-target pair are expensive and impractical. As a result, a more challenging yet practical problem setting, \ie, domain generalization (DG)~\cite{muandet2013domain, li2017deeper} is proposed. Unlike DA, DG aims to train model with multiple source domains that can generalize to arbitrary unseen target domains. To tackle the DG problem, many existing methods utilize adversarial training~\cite{li2018deep, li2018domain, shao2019multi}, meta learning~\cite{Li2018LearningTG, balaji2018metareg, li19l, dou2019domain}, self-supervised learning~\cite{carlucci2019domain} or domain augmentation~\cite{volpi2018generalizing, shankar2018generalizing, zhou2020deep, Zhou2020LearningTG} techniques and have shown promising results. 

\begin{figure}[!t]
    \centering
    \subfigure[]{\label{ori}
        \includegraphics[width=0.15\textwidth]{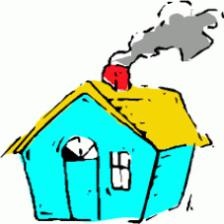}
    }
    \subfigure[]{\label{amp}
        \includegraphics[width=0.15\textwidth]{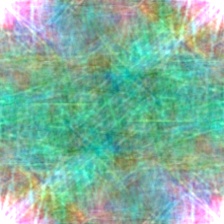}
    }
     \subfigure[]{\label{pha}
        \includegraphics[width=0.15\textwidth]{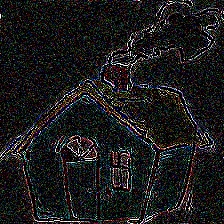}
    }
    \caption{Examples of the amplitude-only and phase-only reconstruction: (a) original image; (b) reconstructed image with amplitude information only by setting the phase component to a constant; (c) reconstructed image with phase information only by setting the amplitude component to a constant.}
    \label{img_recon}
\end{figure}

In this paper, we introduce a novel Fourier-based perspective for DG. Our motivation comes from a well-known property of the Fourier transformation: 
the phase component of Fourier spectrum preserves high-level semantics of the original signal, while the amplitude component contains low-level statistics~\cite{oppenheim1979phase, oppenheim1981importance, piotrowski1982demonstration, hansen2007structural}. For better understanding, we present an example of the images reconstructed from only amplitude information and only phase information, as well as the corresponding original image in Fig.~\ref{img_recon}. As shown in Fig.~\ref{pha}, the phase-only reconstruction reveals the important visual structures, from which one can easily recognize the ``house'' conveyed in the original image. On the other hand, it is hard to tell the exact object from the amplitude-only reconstruction in Fig.~\ref{amp}. 
Based on these observations, Yang \etal ~\cite{yang2020fda} have recently developed a Fourier-based method for DA. They propose a simple image translation strategy by replacing the amplitude spectrum of a source image with that of a random target image. By simply training on the amplitude-transferred source images, their method achieves a remarkable performance. 

\begin{figure*}[!ht]
\begin{center}
\includegraphics[scale=0.65]{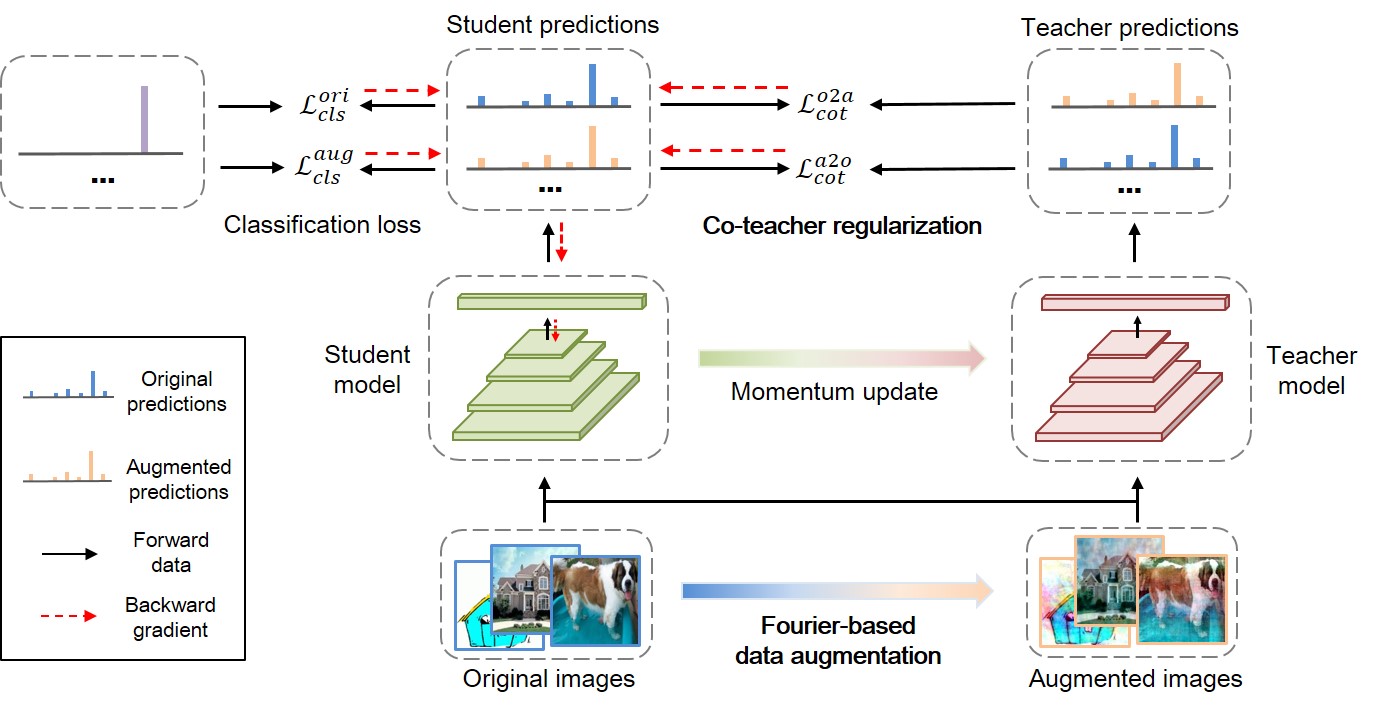}
\end{center}
\caption{The framework of the proposed FACT. Our framework contains two key components, namely Fourier-based data augmentation and co-teacher regularization, which are highlighted in bold. Both the original and augmented data are sent to the student model and a momentum-updated teacher model. The co-teacher regularization then imposes a dual consistency between the predictions from original data and augmented data. Note that the parameters of teacher model are not updated during back propagation.}
\label{framework}
\end{figure*}

Inspired by the above work, we further explore Fourier-based methods for domain generalization and introduce a \textit{Fourier Augmented Co-Teacher} (FACT) framework, which consists of an implicit constraint induced by Fourier-based data augmentation and an explicit constraint in terms of co-teacher regularization, as shown in Fig.~\ref{framework}. 1) \emph{Fourier-based data augmentation}. Since the phase information is known for carrying the essential features to define an object, it is reasonable to assume that by learning more from the phase information, the model can better extract the semantic concepts of different objects that are robust to domain shifts. However, when dealing with DG, we have no access to the target domain, thus the amplitude transfer strategy as~\cite{yang2020fda} is not applicable.  
To overcome this, we propose to augment the training data by distorting the amplitude information while keeping the phase information unchanged. Specifically, a linear interpolation strategy similar to MixUp~\cite{zhang2018mixup} is adopted to generate augmented images. But instead of the whole images, only the amplitude of the images are mixed.
Through this Fourier-based data augmentation, our model can avoid overfitting to low-level statistics carried in the amplitude information, thus pay more attention on the phase information when making decisions.  2) \emph{Co-teacher regularization}. In addition to the above implicit constraint induced by Fourier-based data augmentation, we further introduce an explicit constraint to force the model to maintain the predicted class relationships between the original image and its amplitude-perturbed counterpart. This explicit constraint is designed in a form of dual consistency regularization equipped with a momentum-updated teacher~\cite{tarvainen2017mean}. Through the co-teacher regularization, the model is further constrained to focus on the invariant phase information in order to minimize the prediction discrepancy between original and augmented images. 

We validate the effectiveness of FACT on three domain generalization benchmarks, namely Digits-DG~\cite{zhou2020deep}, PACS~\cite{li2017deeper}, and OfficeHome~\cite{venkateswara2017deep}.
Extensive experimental results have shown that FACT outperforms several state-of-the-arts DG methods with regards to its capability to generalize to unseen domains, indicating that learning more from the phase information does help model generalize better across domains. 
We further carry out detailed ablation studies to show the superiority of our framework design. 
We also conduct an in-depth analysis about the rationales behind our hypothesis and method, which demonstrates that the visual structures in phase information contain rich semantics and our model can learn efficiently from them.

\section{Related Work}

\textbf{Domain generalization}: 
Domain generalization (DG) aims to extract knowledge from multiple source domains so as to generalize well to arbitrary unseen target domains. Early DG studies mainly follow the distribution alignment idea in domain adaptation by learning domain-invariant features via either kernel methods~\cite{muandet2013domain, ghifary2016scatter} or domain-adversarial learning~\cite{li2018deep, li2018domain, shao2019multi}. Later on, Li \etal ~\cite{Li2018LearningTG} propose a meta learning approach that simulates the training strategy of MAML ~\cite{finn17a} by back propagating the second-order gradients calculated on a random meta-test domain split from the source domains at each iteration. Subsequent meta learning-based DG methods utilize a similar strategy to meta-learn a regularizer~\cite{balaji2018metareg}, a feature-critic network~\cite{li19l}, or how to maintain semantic relationships~\cite{dou2019domain}. Another popular way to address DG problem is domain augmentation, which creates samples from fictitious domains via gradient-based adversarial perturbations~\cite{volpi2018generalizing, shankar2018generalizing} or adversarially trained image generators~\cite{zhou2020deep, Zhou2020LearningTG}. Recently, inspired by the robustness of a shape-biased model to out-of-distributions~\cite{geirhos2018imagenettrained}, Shi \etal~\cite{Shi2020InformativeDF} bias their model to shape features by filtering out texture features according to local self-information. Similarly, Carlucci \etal ~\cite{carlucci2019domain} introduce a self-supervised jigsaw task to help the model learn global shape features and Wang \etal~\cite{wang2020learning} further extend this work by incorporating a momentum metric learning scheme. Other DG methods also employ low-rank decomposition~\cite{li2017deeper, Piratla2020EfficientDG} and gradient-guided dropout~\cite{Huang2020SelfChallengingIC}. Different from all the methods above, our work takes a new Fourier-based perspective for DG. By emphasizing the Fourier phase information, our method achieves a remarkable performance compared with current DG methods.

\textbf{The importance of phase information}:
Many early studies~\cite{oppenheim1979phase, oppenheim1981importance, piotrowski1982demonstration, hansen2007structural} have shown that in the Fourier spectrum of signals, the phase component retains most of the high-level semantics in the original signals, while the amplitude component mainly contains low-level statistics. 
Most recently, Yang \etal ~\cite{yang2020fda} introduce the Fourier perspective into domain adaptation. By simply replacing a small area in the centralized amplitude spectrum of a source image with that of a target image, they can generate target-like images for training. Another concurrent work of Yang \etal~\cite{yang2020phase} proposes to keep a phase consistency in source-target image translations, which is shown to be a better choice than the commonly used cycle consistency~\cite{Hoffman_cycada2017} for segmentation tasks under DA scenario. 
Inspired from the above work, we develop a novel Fourier-based framework for DG to encourage our model to focus on the phase information.

\textbf{Consistency regularization}: Consistency regularization (CR) is widely-used in supervised and semi-supervised learning. 
Laine and Aila~\cite{Laine2017TemporalEF} first introduce a consistency loss between outputs from two differently perturbed models. Tarvainen and Valpola~\cite{tarvainen2017mean} further extend this work by using a momentum-updated teacher to provide better targets for consistency alignment. Miyato \etal~\cite{Miyato2019VirtualAT} and Park \etal~\cite{AAAI1816322} develop different techniques by replacing the stochastic perturbations with adversarial ones. Verma \etal~\cite{ijcai2019-504} also demonstrate that a interpolation consistency with MixUp~\cite{zhang2018mixup} samples could be helpful. Some recent work ~\cite{french2017self, shu2018a, wu2020dual} also use consistency regularization in UDA to improve the performance on target domain. In our work, we design a dual-formed consistency loss to further bias our model on the phase information.

\section{Method}
Given a training set of multiple source domains $\mathcal{D}_s = \left\{\mathcal{D}_1, \dots, \mathcal{D}_S\right\}$ with $N_k$ labelled samples $\{(x_{i}^{k}, y_{i}^{k})\}_{i=1}^{N_{k}}$ in the $k$-th domain $\mathcal{D}_k$, where $x_{i}^{k}$ and $y_{i}^{k}$ $\in\{1,\ldots, C\}$ denote the inputs and labels respectively, the goal of domain generalization is to learn a domain-agnostic model $f(\cdot ; \theta)$ on source domains that is expected to perform well on unseen target domains $\mathcal{D}_t$.

Inspired by the semantic-preserving property of Fourier phase component~\cite{oppenheim1979phase, oppenheim1981importance, piotrowski1982demonstration, hansen2007structural}, we assume that models highlight the phase information have better generalization ability across domains. To this end, we design a novel Fourier-based data augmentation strategy by mixing the amplitude information of images. We further add a dual-formed consistency loss, named as co-teacher regularization, to reach consensuses between predictions derived from augmented and original inputs. The consistency loss is implemented with a momentum-updated teacher model to provide better targets for consistency alignment as in~\cite{tarvainen2017mean}. The overall \textit{Fourier-based Augmented Co-Teacher} (FACT) framework is illustrated in Fig.~\ref{framework}. Below we introduce the main components of FACT, \ie, Fourier-based data augmentation and co-teacher regularization.

\subsection{Fourier-based data augmentation}\label{method_fa}
For a single channel image $x$,  its Fourier transformation $\mathcal{F}(x)$ is formulated as:
\begin{equation}
\mathcal{F}(x)(u, v)=\sum_{h=0}^{H-1} \sum_{w=0}^{W-1} x(h, w) e^{-j 2 \pi\left(\displaystyle{\frac{h}{H}} u+\displaystyle{\frac{w}{W}} v\right)}
\end{equation} 
and $\mathcal{F}^{-1}(x)$ defines the inverse Fourier transformation accordingly. 
Both the Fourier transformation and its inverse can be calculated with the FFT algorithm~\cite{nussbaumer1981fast} efficiently. 
The amplitude and phase components are then respectively expressed as:
\begin{equation}\label{abs_pha}
\begin{aligned}
\mathcal{A}(x)(u, v)&=\left[R^{2}(x)(u, v)+I^{2}(x)(u, v)\right]^{1 / 2} \\
\mathcal{P}(x)(u, v)&=\arctan \left[\frac{I(x)(u, v)}{R(x)(u, v)}\right],
\end{aligned}
\end{equation}
where $R(x)$ and $I(x)$ represent the real and imaginary part of $\mathcal{F}(x)$, respectively. For RGB images, the Fourier transformation for each channel  is computed independently to get the corresponding amplitude and phase information.

\begin{figure}[!t]
    \centering
    \subfigure[]{\label{as}
        \includegraphics[width=0.4\textwidth]{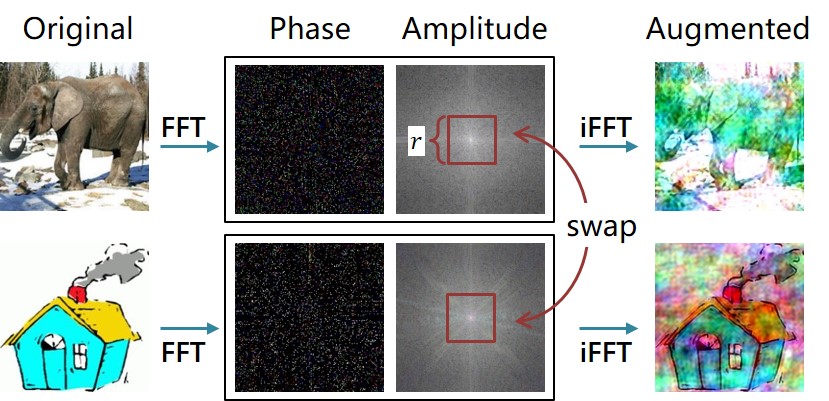}
    }
    \subfigure[]{\label{am}
        \includegraphics[width=0.4\textwidth]{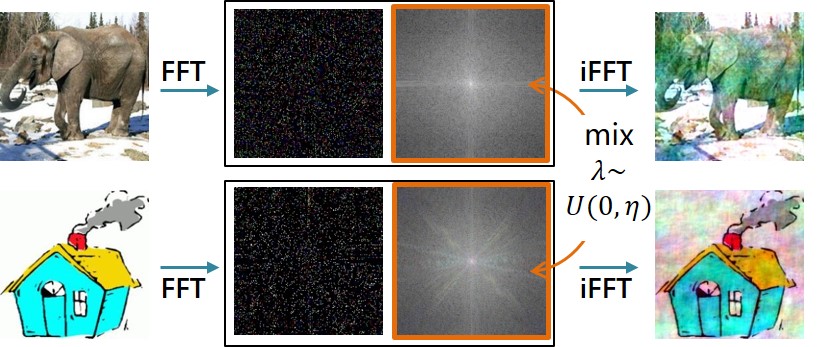}
    }
    \caption{Illustration of (a) AS and (b) AM strategy.}
    \label{augtype}
\end{figure}

With the semantic-preserving property of Fourier phase component, we here attempt to build models that specifically highlight the phase information, which are expected to have better generalization ability across domains. 
To achieve this goal, a natural choice is perturbing the amplitude information in the original images via a certain form of data augmentation.
Inspired from MixUp~\cite{zhang2018mixup}, we design a novel data augmentation strategy by linearly interpolating between the amplitude spectrums of two images from arbitrary source domains:
\begin{equation}
\hat{\mathcal{A}}(x_{i}^{k}) = (1-\lambda)\mathcal{A}(x_{i}^{k}) + \lambda \mathcal{A}(x_{i'}^{k'}),
\end{equation} 
where $\lambda \sim U(0, \eta)$, and the hyperparameter $\eta$ controls the strength of the augmentation. The mixed amplitude spectrum is then combined with the original phase spectrum to form a new Fourier representation:
\begin{equation}
\mathcal{F}(\hat{x}_{i}^{k})(u, v)=\hat{\mathcal{A}}(x_{i}^{k})(u, v) * e^{-j *  \mathcal{P}(x_{i}^{k})(u, v)},
\end{equation}
which is then fed to inverse Fourier transformation to generate the augmented image.  $\hat{x}_{i}^{k}=\mathcal{F}^{-1}[\mathcal{F}(\hat{x}_{i}^{k})(u, v)]$.
This Fourier-based augmentation strategy, named \textit{amplitude mix} (AM) thereafter, is illustrated in Fig.~\ref{augtype}~\subref{am}.

We then feed the augmented images and the original labels to the model for classification. The loss function is formulated as standard cross entropy\footnote{The expectation is omitted for brevity for all the loss functions.}:
\begin{equation}\label{cls_aug}
\mathcal{L}_{cls}^{aug} = - y_{i}^{k} \log \left(\sigma(f(\hat{x}_{i}^{k};\theta))\right)
\end{equation}
where $\sigma$ is the softmax activation. We also use the original images for training, the classification loss $\mathcal{L}_{cls}^{ori}$ can then be defined similarly as Eq.~\ref{cls_aug}.

Note that the AM strategy is essentially different from the spectral transfer operation proposed in~\cite{yang2020fda}. Specifically, the spectral transfer operation aims to adapt low-level statistics from the source domain to the target domain by replacing the low-frequency amplitude information of source images with that of target images. However, in domain generalization, since we have no access to target data, such an adaptive operation is impossible. Nevertheless, we can still swap the amplitude spectrums between two source images directly to create augmented images. This \textit{amplitude swap} (AS) strategy is illustrate in Fig.~\ref{augtype}~\subref{as}. Like~\cite{yang2020fda}, the proportion of the swapped area is controlled by a hyperparameter $r$. However, only swapping a small area of the centralized amplitude spectrum (\ie low-frequency amplitude information) could still cause the model to overfit on the remaining middle-frequency and high-frequency amplitude information, while swapping the whole amplitude spectrum (\ie setting $r=1$) may be too aggressive for the model to learn. On the other hand, the AM strategy perturbs each frequency component in the amplitude spectrum equally and bridge the model to the most aggressively augmented images via linear interpolation. Therefore, the model can efficiently learn from the phase information by comparing between the original and augmented images through AM augmentation.

\subsection{Co-teacher Regularization} \label{method_cot}
The above Fourier-based data augmentation imposes an implicit constraint which requires the model to predict the same object before and after augmentation. However, the categorical relations predicted from original and augmented images with the same phase information may be different. For example, a model may learn from the original images that horses are more similar to giraffes than houses. However, this learned knowledge may conflict to that learned from augmented images due to a different augmentation. To alleviate this kind of disagreement, we add an explicit constraint in the form of a dual consistency loss. 
As suggested in~\cite{tarvainen2017mean}, we use a momentum-updated teacher model to provide better targets for consistency alignment. During training, the teacher model receives parameters from the student model via exponential moving average:
\begin{equation}
\theta_{tea} = m \theta_{tea} + (1-m) \theta_{stu}
\end{equation}
where $m$ is the momentum parameter. Note that no gradient flows through the teacher model during back propagation. To make full use of the knowledge learned from data, we compute the outputs with a softened softmax at a temperature $T$~\cite{hinton2015distilling} for both the teacher and student models.
We then force the model to be consistent between the outputs derived from original images and augmented images:
\begin{align}
\mathcal{L}_{cot}^{a2o} &= \text{KL}(\sigma(f_{stu}(\hat{x}_{i}^{k})/T) || \sigma(f_{tea}(x_{i}^{k})/T)), \label{cot_a2o} \\
\mathcal{L}_{cot}^{o2a} &= \text{KL}(\sigma(f_{stu}(x_{i}^{k})/T) || \sigma(f_{tea}(\hat{x}_{i}^{k})/T)). \label{cot_o2a}
\end{align}

Here we align the student outputs of augmented images to the teacher outputs of original images, as well as the student outputs of original images to the teacher outputs of augmented images. Since the consistency loss takes a dual form and incorporates a momentum teacher, we rename the loss as \emph{co-teacher regularization} for brevity. Through the co-teacher regularization, we want our model to learn equally from the original and augmented images. 
More specifically, the original image and its augmented counterpart can be seen as two views of a same object. When learning from the ``original view'', the student model is not only guided by the ground-truth, but also taught by the teacher model that learns from the ``augmented view''. So is the case when the student model learns from the ``augmented view''. Such a simultaneous co-teaching process enables a comprehensive knowledge sharing between the original and augmented view, and further direct the model to focus on the invariant phase information in order to reach a consistency between the two views.

Combining all the loss functions together, we can get our full objective as:
\begin{equation}\label{total}
\mathcal{L}_{FACT} = \mathcal{L}_{cls}^{ori} + \mathcal{L}_{cls}^{aug} + \beta(\mathcal{L}_{cot}^{a2o} + \mathcal{L}_{cot}^{o2a})
\end{equation}
where $\beta$ controls the trade-off between the classification loss and the co-teacher regularization loss.

\section{Experiment}
In this section, we demonstrate the superiority of our method on several DG benchmarks. We also carry out detailed ablation studies about the impact of different components and augmentation types.

\subsection{Datasets and settings}
We evaluate our method on three conventional DG benchmarks, which cover various recognition scenes. Details of these benchmarks are as follows:
(1) \textbf{Digits-DG}~\cite{zhou2020deep}: a digit recognition benchmark consisted of four classical datasets \textit{MNIST}~\cite{lecun1998gradient}, \textit{MNIST-M}~\cite{ganin2015unsupervised}, \textit{SVHN}~\cite{netzer2011reading}, \text{SYN}~\cite{ganin2015unsupervised}. The four datasets mainly differ in font style, background and image quality. We use the original train-validation split in~\cite{zhou2020deep} with 600 images per class per dataset. 
(2) \textbf{PACS}~\cite{li2017deeper}: an object recognition benchmark including four domains (\textit{photo}, \textit{art-painting}, \textit{cartoon}, \textit{sketch}) with large discrepancy in image styles. It contains seven classes and 9,991 images totally. We use the original train-validation split provided by Li \etal~\cite{li2017deeper}.
(3) \textbf{OfficeHome}~\cite{venkateswara2017deep}: an object recognition benchmark including 15,500 images of 65 classes from four domains (\textit{Art}, \textit{Clipart}, \textit{Product}, \textit{Real-World}). The domain shift mainly comes from image styles and viewpoints, but is much smaller than PACS. Follow~\cite{carlucci2019domain}, we randomly split each domain into $90\%$ for training and $10\%$ for validation.

For all benchmarks, we conduct the leave-one-domain-out evaluation. We train our model on the training splits and select the best model on the validation splits of all source domains. For testing, we evaluate the selected model on all images of the held-out target domain. All the results are reported in terms of classification accuracy and averaged over three runs. 
We employ a vanilla ConvNet trained from a simple aggregation of all source data as our baseline, which is named as DeepAll in the remaining sections.

\begin{table}[!t]
\caption{Leave-one-domain-out results on Digits-DG. The best and second-best results are bolded and underlined respectively.}
  \centering
    \setlength{\tabcolsep}{0.8mm}{\begin{tabular}{l|cccc|c}
    \toprule
    Methods & MNIST & MNIST-M & SVHN & SYN & Avg. \\
    \midrule
    DeepAll~\cite{zhou2020deep} & 95.8 & 58.8 & 61.7 & 78.6 & 73.7 \\
    CCSA~\cite{motiian2017unified} & 95.2 & 58.2 & 65.5 & 79.1 & 74.5 \\
    MMD-AAE~\cite{li2018domain} & 96.5 & 58.4 & 65.0 & 78.4 & 74.6 \\
    CrossGrad~\cite{shankar2018generalizing} & 96.7 & 61.1 & 65.3 & 80.2 & 75.8 \\
    DDAIG~\cite{zhou2020deep} & 96.6 & \underline{64.1} & 68.6 & 81.0 & 77.6 \\
    Jigen~\cite{carlucci2019domain} & 96.5& 61.4 & 63.7 & 74.0 & 73.9 \\
    L2A-OT~\cite{Zhou2020LearningTG} & \underline{96.7} & 63.9 & \underline{68.6} & \underline{83.2} & \underline{78.1} \\
    FACT (\textit{ours}) & \textbf{97.9} & \textbf{65.6} & \textbf{72.4} & \textbf{90.3} & \textbf{81.5} \\
    \bottomrule
    \end{tabular}}
  \label{digitsdg}
\end{table}

\subsection{Implementation details}
We closely follow the implementations of~\cite{carlucci2019domain, zhou2020deep}. Here we briefly introduce the main details for training our model, and more details can be found in the supplementary. 

\textbf{Basic details}: For Digits-DG, we use the same backbone network as~\cite{zhou2020deep, Zhou2020LearningTG}. We train the network from scratch using SGD, batch size of 128 and weight decay of 5e-4 for 50 epochs. The initial learning rate is set to 0.05 and decayed by 0.1 every 20 epochs. For PACS and OfficeHome, we use the ImageNet pretrained ResNet~\cite{he2016deep} as our backbone. We train the network with SGD, batch size of 16 and weight decay of 5e-4 for 50 epochs. The initial learning rate is 0.001 and decayed by 0.1 at 80$\%$ of the total epochs. We also use the standard augmentation protocol as in~\cite{carlucci2019domain},
, which consists of random resized cropping, horizontal flipping and color jittering.

\textbf{Method-specific details}: For all experiments, we set the momentum $m$ for the teacher model to 0.9995 and the temperature $T$ to 10. The weight $\beta$ of the consistency loss is set to 2 for Digits-DG and PACS, and 200 for OfficeHome. 
We also use a sigmoid ramp-up~\cite{tarvainen2017mean} for $\beta$ with a length of 5 epochs. 
The augmentation strength $\eta$ of AM is chosen as 1.0 for Digits-DG and PACS, and 0.2 for OfficeHome.  

\begin{table}[!t]
  \centering
  \caption{Leave-one-domain-out results on PACS. The best and second-best results are bolded and underlined respectively. $\dag$: results are reported based on the best models on test splits.}
    \setlength{\tabcolsep}{1.0mm}{\begin{tabular}{l|cccc|c}
    \toprule
    Methods & Art & Cartoon & Photo & Sketch & Avg. \\
    \midrule
    \multicolumn{6}{c}{\textit{ResNet18}} \\
    \midrule
    DeepAll & 77.63 & 76.77 & 95.85 & 69.50 & 79.94 \\
    MetaReg~\cite{balaji2018metareg} & 83.70& 77.20 & 95.50 & 70.30 & 81.70 \\
    JiGen~\cite{carlucci2019domain} & 79.42 & 75.25 & 96.03 & 71.35 & 80.51 \\
    Epi-FCR~\cite{li2019episodic} & 82.10 & 77.00 & 93.90 & 73.00 & 81.50 \\
    MMLD~\cite{matsuura2020domain}  & 81.28 & 77.16 & 96.09 & 72.29 & 81.83 \\
    DDAIG~\cite{zhou2020deep} & \underline{84.20} & 78.10 & 95.30 & 74.70 & \underline{83.10} \\
    CSD~\cite{Piratla2020EfficientDG} & 78.90 & 75.80 & 94.10 & \underline{76.70} & 81.40 \\
    InfoDrop~\cite{Shi2020InformativeDF} & 80.27 & 76.54 & \underline{96.11} & 76.38 & 82.33 \\
    MASF~\cite{dou2019domain}$^{\dag}$ & 80.29 & 77.17 & 94.99 & 71.69 & 81.04 \\
    L2A-OT~\cite{Zhou2020LearningTG} & 83.30 & 78.20 & \textbf{96.20} & 73.60 & 82.80 \\
    EISNet~\cite{wang2020learning} & 81.89 & 76.44 & 95.93 & 74.33 & 82.15 \\
    RSC~\cite{Huang2020SelfChallengingIC} & 83.43 & 80.31 & 95.99 & 80.85 & 85.15 \\
    RSC (\textit{our imple.})  & 80.55& \textbf{78.60} & 94.43 & 76.02 & 82.40 \\
    FACT (\textit{ours})  & \textbf{85.37} & \underline{78.38} & 95.15 & \textbf{79.15} & \textbf{84.51} \\
    \midrule
    \multicolumn{6}{c}{\textit{ResNet50}} \\
    \midrule
    DeepAll & 84.94 & 76.98 & \textbf{97.64} & 76.75 & 84.08 \\
    MetaReg~\cite{balaji2018metareg} & \underline{87.20} & 79.20 & \underline{97.60} & 70.30 & 83.60 \\
    MASF~\cite{dou2019domain}$^{\dag}$  & 82.89 & 80.49 & 95.01 & 72.29 & 82.67 \\
    EISNet~\cite{wang2020learning} & 86.64 & \underline{81.53} & 97.11 & 78.07 & \underline{85.84} \\
    RSC~\cite{Huang2020SelfChallengingIC} & 87.89 & 82.16 & 97.92 & 83.35 & 87.83 \\
    RSC (\textit{our imple.}) & 83.92 & 79.52 & 95.15 & \underline{82.20} & 85.20 \\
    FACT (\textit{ours}) & \textbf{89.63} & \textbf{81.77} & 96.75 & \textbf{84.46} & \textbf{88.15} \\
    \bottomrule
    \end{tabular}}
  \label{pacs}
\end{table}

\subsection{Evaluation on Digits-DG}

We present the results in Table~\ref{digitsdg}. Among all the competitors, our method achieves the best performance, exceeding the second best method L2A-OT~\cite{Zhou2020LearningTG} by more than 3$\%$ on average. Specifically, on the hardest target domains SVHN and SYN, where involve cluttered digits and low image qualities, our method outperforms L2A-OT with a large margin of 4$\%$ and 7$\%$ respectively. The success of our method indicates that training the model to focus more on the spectral phase information can significantly promote its performance on out-of-domain images.

\subsection{Evaluations on PACS}

The results are shown in Table~\ref{pacs}. It is clearly that our method is among the top performing ones. We notice that the naive DeepAll baseline can get a remarkable accuracy on the photo domain, due to its similarity to the pretrained dataset ImageNet. However, DeepAll performs poorly on the art-painting, cartoon and sketch domains, which bear a larger domain discrepancy. Nevertheless, our FACT can lift the performance of DeepAll by a huge margin of 7.52$\%$ on art-painting, 3.52$\%$ on cartoon and 11.41$\%$ on sketch. 
Meanwhile, the performance of our model on photo domain drops a little. This is reasonable since domains like photo contain complicated and redundant details, and the model may ignore some possibly useful low-level cues by only highlighting the phase information. Nevertheless, our model still achieves a better overall performance.

Compared with the SOTA, our FACT clearly beats the methods based on adversarial data augmentation or meta learning, including the latest MASF~\cite{dou2019domain}, DDAIG~\cite{zhou2020deep} and L2A-OT~\cite{Zhou2020LearningTG}, yet FACT enjoys an efficient training process without any additional adversarial or episodic training steps. The performance of our method also exceeds that of RSC~\cite{Huang2020SelfChallengingIC}, a recent proposed method utilizing a simple yet powerful gradient-guided dropout, by 2.11$\%$ on average\footnote{We rerun the source codes of RSC under the same hyperparameters, but we are unable to reproduce the reported results in original paper~\cite{Huang2020SelfChallengingIC}. We conjecture this may attribute to the difference in hardware environment. For fairness, we compare our method and RSC under our environment.}. All the above comparisons reveal the effectiveness of our method and further demonstrate that the emphasis on phase information improves generalizability across domains.

\begin{table}[!t]
  \centering
  \caption{Leave-one-domain-out results on OfficeHome. The best and second-best results are bolded and underlined respectively.}
    \setlength{\tabcolsep}{1.0mm}{\begin{tabular}{l|cccc|c}
    \toprule
    Methods & Art & Clipart & Product & Real & Avg. \\
    \midrule
    DeepAll & 57.88 & \underline{52.72} & 73.50 & 74.80 & 64.72 \\
    CCSA~\cite{motiian2017unified}  & 59.90 & 49.90 & 74.10 & 75.70 & 64.90 \\
    MMD-AAE~\cite{li2018domain} & 56.50 & 47.30 & 72.10 & 74.80 & 62.70 \\
    CrossGrad~\cite{shankar2018generalizing} & 58.40 & 49.40 & 73.90 & 75.80 & 64.40 \\
    DDAIG~\cite{zhou2020deep} & 59.20 & 52.30 & \underline{74.60} & 76.00 & 65.50 \\
    L2A-OT~\cite{Zhou2020LearningTG} & \textbf{60.60} & 50.10 & \textbf{74.80} & \textbf{77.00} & \underline{65.60} \\
    Jigen~\cite{carlucci2019domain} & 53.04 & 47.51 & 71.47 & 72.79 & 61.20 \\
    RSC~\cite{Huang2020SelfChallengingIC}   & 58.42 & 47.90 & 71.63 & 74.54 & 63.12 \\
    Jigen (\textit{our imple.}) & 57.95 & 49.21 & 72.61 & 74.90 & 63.67 \\
    RSC (\textit{our imple.}) & 57.67 & 48.48 & 72.62 & 74.16 & 63.23 \\
    FACT (\textit{ours}) & \underline{60.34} & \textbf{54.85} & 74.48 & \underline{76.55} & \textbf{66.56} \\
    \bottomrule
    \end{tabular}}
  \label{officehome}
\end{table}

\subsection{Evaluations on OfficeHome}

We report the results in Table~\ref{officehome}. Due to a relatively smaller domain discrepancy and the similarity to the pretrained dataset ImageNet, DeepAll acts as a strong baseline on OfficeHome. Many previous DG methods, such as CCSA~\cite{motiian2017unified}, MMD-AAE~\cite{li2018domain}, CrossGrad~\cite{shankar2018generalizing} and Jigen~\cite{carlucci2019domain}, can not improve much over the simple DeepAll baseline. Nevertheless, our FACT achieves a consistent improvement over DeepAll on all the held-out domains. Moreover, FACT also the surpasses the latest DDAIG~\cite{zhou2020deep} and L2A-OT~\cite{zhou2020deep} in terms of average performance. This again justifies the superiority of our method. 

\subsection{Ablation studies}
\textbf{Impact of different components}:
We conduct an extensive ablation study to investigate the role of each component in our FACT model in Table~\ref{ablation_component}. Starting from baseline, model A is trained with the AM augmentation only and already works better than the strongest competitor DDAIG~\cite{zhou2020deep} in Table~\ref{pacs}. Based on model A, we add a vanilla dual-formed consistency loss to obtain model B, which improves over model A slightly. Further incorporating the momentum teacher results in our FACT, which performs best in all variants. This indicates the importance of the momentum teacher that provides better targets for consistency loss. We also create a model C by excluding the AM augmentation from FACT and the performance drops a lot, showing that the Fourier-based data augmentation plays a crucial role. We further validate the necessity of the dual form in co-teacher regularization by using only $\mathcal{L}_{cot}^{a2o}$ or $\mathcal{L}_{cot}^{o2a}$, and resulting in model D and E respectively. As in Table~\ref{ablation_component}, neither model D or E outperforms the full FACT, suggesting the effectiveness of incorporating both original-to-augmented and augmented-to-original consistency alignment through co-teacher regularization.

\begin{table*}[htbp]
  \centering
  \caption{Ablation studies on different components of our method on the PACS dataset with ResNet18.}
    \begin{tabular}{l|cccc|cccc|c}
    \toprule
    Method & AM & $\mathcal{L}_{cot}^{a2o}$ & $\mathcal{L}_{cot}^{o2a}$ & Teacher & Art & Cartoon & Photo & Sketch & Avg. \\
    \midrule
    Baseline & \textbf{-} & \textbf{-} & \textbf{-} & \textbf{-} & 77.63±0.84 & 76.77±0.33 & \textbf{95.85±0.20} & 69.50±1.26 & 79.94 \\
    \midrule
    Model A & \checkmark & \textbf{-} & \textbf{-} & \textbf{-} & 83.90±0.50 & 76.95±0.45 & 95.55±0.12 & 77.36±0.71 & 83.44 \\
    Model B & \checkmark & \checkmark & \checkmark & \textbf{-} & 83.71±0.30 & 77.84±0.49 & 94.73±0.12 & 78.55±0.46 & 83.71 \\
    Model C & \textbf{-} & \checkmark & \checkmark & \checkmark & 82.68±0.44 & 78.06±0.39 & 95.35±0.44 & 74.76±0.67 & 82.71 \\
    Model D & \checkmark & \checkmark & \textbf{-} & \checkmark & 83.97±0.77 & 77.04±0.86 & 94.59±0.03 & 79.08±0.56 & 83.67 \\
    Model E & \checkmark & \textbf{-} & \checkmark & \checkmark & 84.07±0.43 & 77.70±0.65 & 95.28±0.34 & 78.29±0.61 & 83.84 \\
    \midrule
    FACT & \checkmark & \checkmark & \checkmark & \checkmark & \textbf{85.37±0.29} & \textbf{78.38±0.29} & 95.15±0.26 & \textbf{79.15±0.69} & \textbf{84.51} \\
    \bottomrule
    \end{tabular}
  \label{ablation_component}
\end{table*}

\vspace{3pt}
\textbf{Other choices of Fourier-based data augmentation}: Next we show the advantages of our AM augmentation over its alternatives in Table~\ref{ablation_aug}. Specifically, we compare the AM strategy with the AS strategy which is mentioned in Sec.~\ref{method_fa}. Follow~\cite{yang2020fda}, we first choose to swap only a small area in the centralized amplitude spectrum by setting ratio $r=0.09$ , and the resulting strategy is called AS-partial. As in Table~\ref{ablation_aug}, the performance of AS-partial is inferior to that of AM. This is reasonable, as choosing a small value of $r$ in AS only perturbs the low-frequency components in the amplitude spectrum, while the model still has risks to overfit on the remaining amplitude information. Nevertheless, we can still perturb all frequency components with AS by setting $r=1.0$, and the resulting strategy is called AS-full. This choice brings some improvement but is still worse than AM. We attribute this to the negative effect caused by directly swapping the entire amplitude spectrums of two images, which may be too aggressive for the model to learn. 

\begin{table}[!t]
  \centering
  \caption{Ablation studies of different choices of the Fourier data augmentation on the PACS dataset with ResNet18.}
    \scalebox{0.95}{\setlength{\tabcolsep}{2mm}{\begin{tabular}{l|cccc|c}
    \toprule
    \makecell[c]{Methods} & Art & Cartoon & Photo & Sketch & Avg. \\
    \midrule
    \multicolumn{6}{c}{\textit{DeepAll with}} \\ \midrule
    \hspace{0.5em}AS-partial & 82.00 & 76.19 & 93.89 & 77.27 & 82.34 \\
    \hspace{0.5em}AS-full & 83.50 & 76.07 & 94.49 & 77.13 & 82.80 \\
    \hspace{0.5em}AM & \textbf{83.90} & \textbf{76.95} & \textbf{95.55} & \textbf{77.36} & \textbf{83.44} \\
    \midrule
    \multicolumn{6}{c}{\textit{FACT with}} \\ \midrule
    \hspace{0.5em}AS-partial & 81.61 & 76.95 & 93.83 & 78.30 & 82.67 \\
    \hspace{0.5em}AS-full & 83.46 & 77.37 & 94.10 & 78.63 & 83.39 \\
    \hspace{0.5em}AM & \textbf{85.37} & \textbf{78.38} & \textbf{95.15} & \textbf{79.15} & \textbf{84.51} \\
    \bottomrule
    \end{tabular}}}
  \label{ablation_aug}
\end{table}

\begin{table}[!t]
\centering
\caption{The performance changes of training with phase-only reconstructed images and amplitude-only reconstructed images when compared with original images. The values greater than zero (meaning an improvement) are in bold.}
\scalebox{0.9}{\setlength{\tabcolsep}{1.5mm}{\begin{tabular}{c|ccccc}
\toprule
Data & \multicolumn{1}{c|}{\diagbox{Train}{Test}} & Photo  & Art    & Cartoon & Sketch \\ \midrule
\multirow{4}{*}{\begin{tabular}[c]{@{}c@{}}Phase\\ only\end{tabular}} & \multicolumn{1}{c|}{Photo}   & -4.68  & \textbf{3.16} & \textbf{4.07} & \textbf{2.38}   \\
 & \multicolumn{1}{c|}{Art}     & -5.35  & \textbf{1.28}   & \textbf{5.97}    & \textbf{15.87}  \\
 & \multicolumn{1}{c|}{Cartoon} & -11.53 & \textbf{0.29}   & -4.08   & \textbf{18.55}  \\
 & \multicolumn{1}{c|}{Sketch}  & \textbf{10.66}  & \textbf{14.56}  & \textbf{21.26}   & -1.09  \\ \midrule
\multirow{4}{*}{\begin{tabular}[c]{@{}c@{}}Amplitude\\ only\end{tabular}} & \multicolumn{1}{c|}{Photo}   & -14.03 & -4.15  & -4.41   & -0.08  \\
 & \multicolumn{1}{c|}{Art}     & -18.40 & -21.96 & -5.59   & -10.72 \\
 & \multicolumn{1}{c|}{Cartoon} & -13.95 & -7.48  & -15.89  & \textbf{1.36}   \\
 & \multicolumn{1}{c|}{Sketch}  & -4.79  & -0.73  & -1.99   & -13.99 \\ \bottomrule
\end{tabular}}}
\label{dis_single}
\end{table}

\section{Discussion}

\textbf{Phase information contains meaningful semantics and helps generalization.} In previous sections, we have seen that by learning more from the phase information, the model can generalize well on unseen domains. Here we further verify the importance of phase information through single domain evaluations on PACS. 
Specifically, we first generate phase-only reconstructed images by setting the amplitude component as a constant, and so is the amplitude-only reconstructed images. Then we train three models on original images, phase-only images and amplitude-only images respectively, and compare their performance in Table~\ref{dis_single}. For fair comparison, all the models here are trained from scratch without ImageNet pretraining.
As shown in Table~\ref{dis_single}, the model trained with phase-only images performs better, or at least comparable with the baseline trained with original images in 11 out of 16 cases. This indicates that the phase information does contain useful semantics to help the model generalize to unseen domains. On the other hand, the model trained with amplitude-only images suffers from large performance degradation in almost all the cases, suggesting that the amplitude information hardly contains any meaningful semantics. Another interesting finding is that the model trained with phase-only images has some performance drops when generalizing to photo domain from cartoon and art domain. We conjecture that a desirable performance on the photo domain may also require the presence of amplitude information. Therefore, in our FACT framework, we do not completely eliminate the amplitude information, but instead shift model's attention to the phase information in an amplitude-perturbation fashion. 
It worths nothing that the performance of the model trained with phase-only images also declines in some in-domain generalization cases. This is reasonable considering the loss of amplitude information. 

\vspace{3pt}
\textbf{Amplitude perturbation constrains the model to focus more on phase information.} Our Fourier-based data augmentation are implemented via perturbing the amplitude information. Here we present a brief theoretical analysis to demonstrate that amplitude perturbation does make the model to focus more on phase information. For simplicity, we consider the case of a linear softmax classifier together with a feature extractor $\mathbf{h}$. Suppose the distribution of Fourier-based data augmentation is $g \sim \mathcal{G}$, and the risk of training on the augmented data is:
\begin{equation}
\hat{R}_{\text {aug }}=\frac{1}{N} \sum_{n=1}^{N} \mathbb{E}_{g \sim \mathcal{G}}\left[\ell\left(\mathbf{W}^{\top} \mathbf{h}\left(g\left(x\right)\right), y\right)\right]
\end{equation}
Similar as in~\cite{Dao2019AKT, He2019DataAR}, we expand $\hat{R}_{\text {aug }}$ with Taylor expansion:
\begin{equation}
\label{taylor}
\begin{array}{l}
\mathbb{E}_{g \sim \mathcal{G}}\left[\ell\left(\mathbf{W}^{\top} \mathbf{h}(g(x)), y\right)\right]= \\
\qquad \qquad \ell\left(\mathbf{W}^{\top} \overline{\mathbf{h}}, y\right)+\frac{1}{2} \mathbb{E}_{g \sim \mathcal{G}}\left[\Delta^{\top} \mathbf{H}(\tau, y) \Delta\right]
\end{array}
\end{equation}
where $\overline{\mathbf{h}}=\mathbb{E}_{g \sim \mathcal{G}}[\mathbf{h}(g(x))]$, $\Delta=\mathbf{W}^{\top}(\overline{\mathbf{h}}-\mathbf{h}(g(x)))$ and $\mathbf{H}$ is the Hessian matrix. For cross-entropy loss with softmax, $\mathbf{H}$ is semi-definite and independent of $y$. Then, minimizing the second-order term in Eq.~\ref{taylor} requires that for some feature $h_d$, if its variance $h_{d}(g(x))$ is large, the weight $w_{i, d}$ will approach 0. Suppose that the features induced from phase information and amplitude information is $h_p$ and $h_a$ respectively, since we only perturb the amplitude information and keep the phase information unchanged, it is reasonable that:
\begin{equation}
\left\{
\begin{array}{l}
\left|h_{p}(g(x))-h_{p}(x)\right|<\zeta \\
\left|h_{a}(g(x))-h_{a}(x)\right|>\epsilon
\end{array}
\right.
\end{equation}
where $\zeta>0$ is a small value, and $\epsilon \geq \zeta$. Therefore, minimizing $\hat{R}_{\text {aug }}$ restricts $w_{i, a} \rightarrow 0$ for those features $h_a$ derived from the amplitude information. As a result, the classifier would pay more attention to the features $h_p$ that derived from the phase information when making decisions.

\begin{table}[!h]
\centering
\caption{The cosine similarities between the phase-only reconstructed images (P) and the edge images detected by Sobel operator (S) and Laplacian operator (L).}
\scalebox{0.9}{\setlength{\tabcolsep}{6.0mm}{\begin{tabular}{l|lll}
\toprule
\multicolumn{1}{c|}{} & \multicolumn{1}{c}{S\&P} & \multicolumn{1}{c}{L\&P} & \multicolumn{1}{c}{S\&L} \\ \midrule
Similiarity           & 0.790                    & 0.914                    & 0.873                    \\ \bottomrule
\end{tabular}}}
\label{dis_edge}
\end{table}

\textbf{Why does the phase information provides model with meaningful semantics?} The answer may be that the phase information records the ``location'' of events~\cite{oppenheim1981importance} (or small local structures) and reveals the spatial relationships between them within a given image~\cite{hansen2007structural}. The model can then aggregate these cues to gain a correct knowledge about the objects conveyed in the image. A similar mechanism can also be found in human vision systems~\cite{hansen2007structural}. We also notice that the phase-only reconstructed image in Fig.~\ref{pha} mainly preserves the contours and edges of the original image. For further investigation, we compute the cosine similarities between phase-only images and edge-detected images produced by Sobel or Laplacian operators in Table~\ref{dis_edge}. As we can seen, the similarity scores between phase-only images and edge-detected images are very high, implying that the visual structures such as edges and contours are carried in the phase information. Since the visual structures are the keys to describe different objects, regardless of the underlying domain distributions, learning from such information can facilitate model to extract high-level semantics. 

\section{Conclusions}
In this paper, we introduce a novel perspective based on Fourier transformation into domain generalization. The main idea is that learning more from the spectral phase information can help the model capture domain-invariant semantic concepts. We then propose a framework composed of an implicit constraint induced by Fourier-based data augmentation and an explicit constraint induced by co-teacher regularization. Extensive experiments on three benchmarks demonstrate that our method is able to achieve state-of-the-art performance for domain generalization.
Furthermore, we conduct an in-depth analysis about the mechanisms and rationales behind our method, which gives us a better knowledge about why focusing on the phase information can help domain generalization.
Considering the mainstream of related work is still domain-adversarial learning or meta learning, we hope our work can shed some lights into the community.

\vspace{0.5cm}
\noindent \textbf{Acknowledgement}: This work is supported by the National Key Research and Development Program of China (No. 2019YFB1804304), SHEITC (No. 2018-RGZN-02046), 111 plan (No. BP0719010),  and STCSM (No. 18DZ2270700), and State Key Laboratory of UHD Video and Audio Production and Presentation.

{\small
\bibliographystyle{ieee_fullname}
\bibliography{egbib}
}

\clearpage

\appendix
\input{supp_input.tex}

\end{document}

%% file: supp_input.tex



\section{More examples of amplitude-only and phase-only reconstruction}
\begin{figure*}[b]
\begin{center}
\includegraphics[scale=0.65]{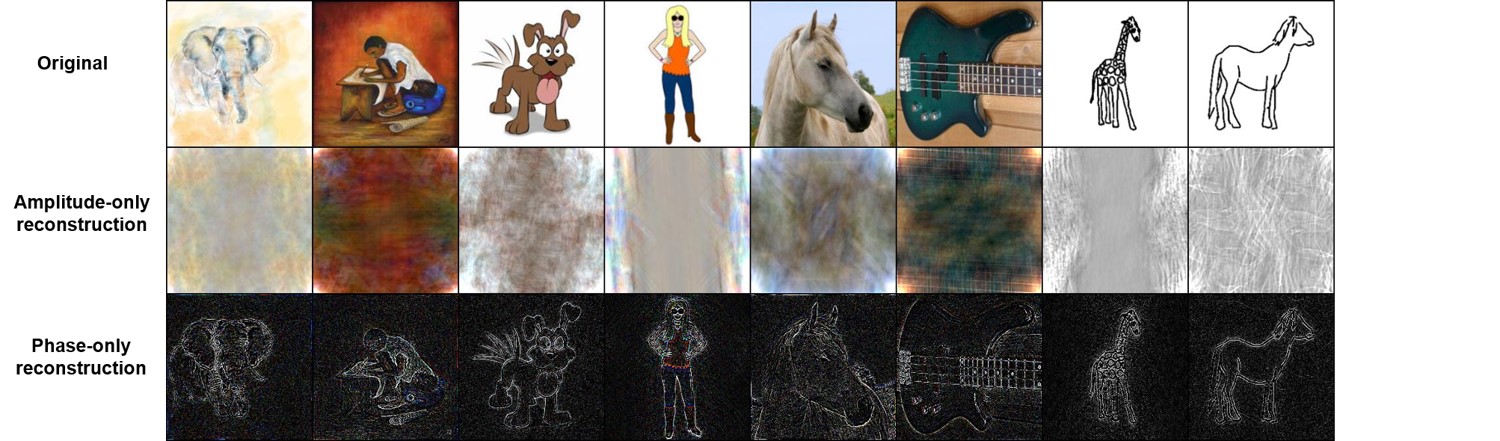}
\end{center}
\caption{Examples of amplitude-only and phase-only reconstruction. }
\label{demo_recon}
\end{figure*}

We present more examples of amplitude-only reconstructed images, phase-only reconstructed images, as well as their corresponding original images in Fig.~\ref{demo_recon}. As we can see, the general visual structures of different objects are preserved in the phase-only reconstructed images, while the amplitude-only reconstructed images mainly contain low-level statistics without clear semantic meanings.  

\section{Implementation details}

\subsection{Experiments on three DG benchmarks}

\textbf{Network details}: We closely follow the implementations of~\cite{carlucci2019domain, zhou2020deep}. For Digits-DG, we use the same backbone network as~\cite{zhou2020deep}.
For PACS and OfficeHome, we use ImageNet pre-trained ResNet18 and ResNet50 as the backbone. 

\textbf{Optimization details}: For all the datasets, we train our network using the nesterov-momentum SGD with a momentum of 0.9 and weight decay of 5e-4. For Digits-DG and PACS, we train the model for 50 epochs. For OfficeHome, we train the model for 30 epochs. The initial learning rate for Digits-DG is 0.05 and decayed by 0.1 every 20 epochs. For PACS and OfficeHome, the initial learning rate is 0.001 and decayed by 0.1 at 80$\%$ of the total epochs. The batch size is set to 128 for Digits-DG and 16 for PACS and OfficeHome. 

\textbf{Data Augmentation details}: The term ``augmentation'' here refers to the typical data augmentation techniques. For Digits-DG, we only use simple augmentations composed of random flipping and resizing. The input image size is $32 \times 32$. For PACS and OfficeHome, we use the standard augmentation protocol as in ~\cite{carlucci2019domain}, which consists of randomly cropping the images to retain between 80$\%$ to 100$\%$, randomly applied horizontal flipping and randomly color jittering with magnitude of 0.4. The input image size is $224 \times 224$.

\textbf{Model-specific details}: For all experiments, we set the momentum $m$ for the teacher model to 0.9995 and the temperature $T$ to 10. The weight $\beta$ of the consistency loss is set to 2 for Digits-DG and PACS, and 200 for OfficeHome. We also use a sigmoid ramp-up~\cite{tarvainen2017mean} for $\beta$ with a length of 5 epochs. The augmentation strength of AM is chosen as 1.0 for Digits-DG and PACS, and 0.2 for OfficeHome. For the convenience of applying amplitude mixing, when an original image is generated, we sample another image from the whole dataset, and then mixing the amplitude spectrums of these two images to produce two augmented counterparts. We then pass these two original images as well as their augmented counterparts to the model. Therefore, within a single iteration, the number of input images is $4\times$ batch size.

\begin{table*}[!htbp]
  \centering
  \caption{Leave-one-domain-out results on Digits-DG.}
    \begin{tabular}{l|cccc|c}
    \toprule
    Methods & MNIST & MNIST-M & SVHN & SYN & Avg. \\
    \midrule
    Jigen~\cite{carlucci2019domain} & 96.5& 61.4 & 63.7 & 74.0 & 73.9 \\
    L2A-OT~\cite{Zhou2020LearningTG} & 96.7 & 63.9 & 68.6 & 83.2 & 78.1 \\
    \midrule
    DeepAll~\cite{zhou2020deep} & 95.8±0.3 & 58.8±0.5 & 61.7±0.5 & 78.6±0.6 & 73.7 \\
    CCSA~\cite{motiian2017unified} & 95.2±0.2 & 58.2±0.6 & 65.5±0.2 & 79.1±0.8 & 74.5 \\
    MMD-AAE~\cite{li2018domain} & 96.5±0.1 & 58.4±0.1 & 65.0±0.1 & 78.4±0.2 & 74.6 \\
    CrossGrad~\cite{shankar2018generalizing} & 96.7±0.1 & 61.1±0.5 & 65.3±0.5 & 80.2±0.2 & 75.8 \\
    DDAIG~\cite{zhou2020deep} & 96.6±0.2 & 64.1±0.4 & 68.6±0.6 & 81.0±0.5 & 77.6 \\
    FACT (\textit{ours}) & \textbf{97.9±0.2} & \textbf{65.6±0.4} & \textbf{72.4±0.7} & \textbf{90.3±0.1} & \textbf{81.5} \\
    \bottomrule
    \end{tabular}
  \label{digitsdg_full}
\end{table*}

\begin{table*}[!htbp]
  \centering
  \caption{Leave-one-domain-out results on OfficeHome.}
    \begin{tabular}{l|cccc|c}
    \toprule
    Methods & Art & Clipart & Product & Real & Avg. \\
    \midrule
    Jigen~\cite{carlucci2019domain} & 53.04 & 47.51 & 71.47 & 72.79 & 61.20 \\
    RSC~\cite{Huang2020SelfChallengingIC}   & 58.42 & 47.90 & 71.63 & 74.54 & 63.12 \\
    L2A-OT~\cite{zhou2020deep} & 60.60 & 50.10 & 74.80 & 77.00 & 65.60 \\
    \midrule
    DeepAll & 57.88±0.20 & 52.72±0.50 & 73.50±0.30 & 74.80±0.10 & 64.72 \\
    CCSA~\cite{motiian2017unified}  & 59.90±0.30 & 49.90±0.40 & 74.10±0.20 & 75.70±0.20 & 64.90 \\
    MMD-AAE~\cite{li2018domain} & 56.50±0.40 & 47.30±0.30 & 72.10±0.30 & 74.80±0.20 & 62.70 \\
    CrossGrad~\cite{shankar2018generalizing} & 58.40±0.70 & 49.40±0.40 & 73.90±0.20 & 75.80±0.10 & 64.40 \\
    DDAIG~\cite{zhou2020deep} & 59.20±0.10 & 52.30±0.30 & \textbf{74.60±0.30} & 76.00±0.10 & 65.50 \\
    Jigen (\textit{our imple.}) & 57.95±0.62 & 49.21±0.35 & 72.61±0.45 & 74.90±0.25 & 63.67 \\
    RSC (\textit{our imple.}) & 57.67±0.51 & 48.48±0.44 & 72.62±0.31 & 74.16±0.48 & 63.23  \\
    FACT (\textit(ours)) & \textbf{60.34±0.11} & \textbf{54.85±0.37} & 74.48±0.13 & \textbf{76.55±0.10} & \textbf{66.56} \\
    \bottomrule
    \end{tabular}
  \label{officehome_full}
\end{table*}

\begin{table*}[htbp]
  \centering
  \caption{Leave-one-domain-out results on PACS. $\dag$: results are reported based on the best models on test splits.}
    \setlength{\tabcolsep}{4.5mm}{\begin{tabular}{lccccc}
    \toprule
    Methods & Art & Cartoon & Photo & Sketch & Avg. \\
    \toprule
    \multicolumn{6}{c}{\textit{ResNet18}} \\
    \toprule
    JiGen~\cite{carlucci2019domain} & 79.42 & 75.25 & 96.03 & 71.35 & 80.51 \\
    Epi-FCR~\cite{li2019episodic} & 82.10 & 77.00 & 93.90 & 73.00 & 81.50 \\
    MMLD~\cite{matsuura2020domain}  & 81.28 & 77.16 & 96.09 & 72.29 & 81.83 \\
    InfoDrop~\cite{Shi2020InformativeDF} & 80.27 & 76.54 & 96.11 & 76.38 & 82.33 \\
    L2A-OT~\cite{Zhou2020LearningTG} & 83.30 & 78.20 & 96.20 & 73.60 & 82.80 \\
    RSC~\cite{Huang2020SelfChallengingIC} & 83.43 & 80.31 & 95.99 & 80.85 & 85.15 \\
    \midrule
    DeepAll & 77.63±0.84 & 76.77±0.33 & 95.85±0.20 & 69.50±1.26 & 79.94 \\
    MetaReg~\cite{balaji2018metareg} & 83.70±0.19 & 77.20±0.31 & 95.50±0.24 & 70.30±0.28 & 81.70 \\
    DDAIG~\cite{zhou2020deep} & 84.20±0.30 & 78.10±0.60 & 95.30±0.40 & 74.70±0.80 & 83.10 \\
    CSD~\cite{Piratla2020EfficientDG} & 78.90±1.10 & 75.80±1.00 & 94.10±0.20 & 76.70±1.20 & 81.40 \\
    EISNet~\cite{wang2020learning} & 81.89±0.88 & 76.44±0.31 & \textbf{95.93±0.06} & 74.33±1.37 & 82.15 \\
    RSC (\textit{our imple.})   & 80.55±0.78 & \textbf{78.60±0.38} & 94.43±0.01 & 76.02±1.68 & 82.40 \\
    FACT (\textit{ours})  & \textbf{85.37±0.29} & 78.38±0.29 & 95.15±0.26 & \textbf{79.15±0.69} & \textbf{84.51} \\
    \midrule
    MASF~\cite{dou2019domain}$^{\dag}$ & 80.29±0.18 & 77.17±0.08 & 94.99±0.09 & 71.69±0.22 & 81.04 \\
    FACT (\textit{ours})$^{\dag}$  & \textbf{85.90±0.27} & \textbf{79.35±0.03} & \textbf{96.61±0.17} & \textbf{80.89±0.26} & \textbf{85.69} \\
    \toprule
    \multicolumn{6}{c}{\textit{ResNet50}} \\
    \toprule
    RSC~\cite{Huang2020SelfChallengingIC} & 87.89 & 82.16 & 97.92 & 83.35 & 87.83 \\
    \midrule
    DeepAll~\cite{Huang2020SelfChallengingIC} & 84.94±0.66 & 76.98±1.13 & \textbf{97.64±0.10} & 76.75±0.41 & 84.08 \\
    MetaReg~\cite{balaji2018metareg} & 87.20±0.13 & 79.20±0.27 & 97.60±0.31 & 70.30±0.18 & 83.60 \\
    EISNet~\cite{wang2020learning} & 86.64±1.41 & 81.53±0.64 & 97.11±0.40 & 78.07±1.43 & 85.84 \\
    RSC (\textit{our imple.}) & 83.92±1.02 & 79.52±2.17 & 95.15±0.10 & 82.20±1.28 & 85.20 \\
    FACT (\textit{ours}) & \textbf{89.63±0.51} & \textbf{81.77±0.19} & 96.75±0.10 & \textbf{84.46±0.78} & \textbf{88.15} \\
    \midrule
    MASF~\cite{dou2019domain}$^{\dag}$  & 82.89±0.16 & 80.49±0.21 & 95.01±0.10 & 72.29±0.15 & 82.67 \\
    FACT (\textit{ours})$^{\dag}$ & \textbf{90.89±0.19} & \textbf{83.65±0.12} & \textbf{97.78±0.05} & \textbf{86.17±0.14} & \textbf{89.62} \\
    \bottomrule
    \end{tabular}}
  \label{pacs_full}
\end{table*}

\subsection{Single domain evaluations on PACS}

Here we present the experimental details about the single domain evaluations in the \textbf{Discussion} of main paper. Specifically, we train ResNet18 using the original images, phase-only reconstructed images and amplitude-only reconstructed images, respectively, on the training splits from a single domain, and then evaluate on the validation splits of all domains. The phase-only reconstructed images are generated by setting the amplitude component as a constant of 20000 when reconstructing the images via inverse FFT. Through this way, the amplitude information is eliminated in the phase-only images so that the model only relies on the phase information for classification. Similar operations are applied to get the amplitude-only reconstructed images. Since the distributions of phase-only and amplitude-only images differ drastically from the original images, we train the networks from scratch in order to remove the impacts of ImageNet pre-training. This ensures a fair comparison between the performances of the original images, phase-only reconstructed images and amplitude-only reconstructed images. Other basic settings are kept the same with the above DG experiments on PACS.

\section{Complete results on DG benchmarks}
We present the complete results in the form of mean±std on Digits-DG, PACS, OfficeHome in Table~\ref{digitsdg_full},  Table~\ref{officehome_full}, Table~\ref{pacs_full}, respectively. Note that unlike most previous work, Dou \etal directly report the best accuracy on the target domain~\cite{dou2019domain}. For fair comparison, we also report our results under this protocol in Table~\ref{pacs_full}. 

\section{Additional results of AlexNet}
To further verify the flexibility of our framework with different backbone networks, we experiment on PACS by incorporating AlexNet into our FACT framework. Specifically, we use the Caffe-version of ImageNet pretrained AlexNet\footnote{\url{https://drive.google.com/file/d/1wUJTH1Joq2KAgrUDeKJghP1Wf7Q9w4z-/view}}. We train the network with nesterov-momentum SGD, batch size of 32 and weight decay of 5e-4 for 50 epochs. The initial learning rate is 0.001 and decayed by 0.1 at 80$\%$ of the total epochs. We also use the standard augmentation protocol as in~\cite{carlucci2019domain}, which consists of random resized cropping, horizontal flipping and color jittering. We set the momentum $m$ for the teacher model to 0.9995 and the temperature $T$ to 10. The weight $\beta$ of the consistency loss is set to 2. We also use a sigmoid ramp-up~\cite{tarvainen2017mean} for $\beta$ with a length of 5 epochs. The augmentation strength of AM is chosen as 1.0. 

\begin{table*}[!htbp]
\centering
\caption{Leave-one-domain-out results on PACS with AlexNet as backbone. $\dag$: results are reported based on the best models on test splits.}
\setlength{\tabcolsep}{6.5mm}{\begin{tabular}{l|cccc|c}
\toprule
Methods & \multicolumn{1}{c}{Art}        & \multicolumn{1}{c}{Cartoon}    & \multicolumn{1}{c}{Photo}      & Sketch                          & Avg.                      \\ \midrule
DeepAll~\cite{carlucci2019domain}                      & 66.68                          & 69.41                          & 89.98                          & 60.02                           & 71.52                     \\
JiGen~\cite{carlucci2019domain}                        & 67.63                          & 71.71                          & 89.00                          & 65.18                           & 73.38                     \\
Epi-FCR~\cite{li2019episodic}                      & 64.70                          & 72.30                          & 86.10                          & 65.00                           & 72.00                     \\
MMLD~\cite{matsuura2020domain}                         & 69.27                          & 72.83                          & 88.98                          & 66.44                           & 74.38                     \\
RSC~\cite{Huang2020SelfChallengingIC}                          & 71.62                          & 75.11                          & 90.88                          & 66.62                           & 76.05                     \\ \midrule
DeepAll & 65.60±0.34 & 70.88±0.29 & 87.16±0.19 & 66.43±0.68 & 72.52 \\
EISNet~\cite{wang2020learning}                       & \multicolumn{1}{c}{70.38±0.37} & \multicolumn{1}{c}{71.59±1.32} & \multicolumn{1}{c}{91.20±0.00} & \multicolumn{1}{c|}{70.25±1.36} & \multicolumn{1}{c}{75.86} \\
MetaVIB~\cite{Du2020LearningTL}                     & \multicolumn{1}{c}{71.94±0.34} & \multicolumn{1}{l}{\textbf{73.17±0.21}} & \multicolumn{1}{c}{\textbf{91.93±0.23}} & \multicolumn{1}{l|}{65.94±0.24} & \multicolumn{1}{c}{75.74} \\
FACT (\textit{Ours})                         & \multicolumn{1}{c}{\textbf{75.50±0.52}} & \multicolumn{1}{c}{71.16±0.24} & \multicolumn{1}{c}{89.10±0.20} & \multicolumn{1}{c|}{\textbf{71.65±0.39}} & \multicolumn{1}{c}{\textbf{76.85}} \\ \midrule
MASF~\cite{dou2019domain}$^{\dag}$                         & 70.35±0.33                     & 72.46±0.19                     & \textbf{90.68±0.12}            & 67.33±0.12                      & 75.21                     \\
FACT (\textit{Ours})$^{\dag}$                        & \textbf{76.46±0.28}            & \textbf{72.57±0.39}            & 90.24±0.26                     & \textbf{73.56±0.08}             & \textbf{78.21}            \\ \bottomrule
\end{tabular}}
\label{pacs_alexnet}
\end{table*}

The results are presented in Table~\ref{pacs_alexnet}, which have shown that FACT with AlexNet as backbone is still able to outperform the state-of-the-arts, by exceeding both EISNet~\cite{wang2020learning} and MetaVIB~\cite{Du2020LearningTL} by around 1$\%$ in terms of  the average performance. The largest performance gain of FACT comes from the generalization tasks on art and sketch domain, both of which bear a large distribution shift from the pre-training domain of ImageNet. This demonstrates the effectiveness of our method when generalizing to unknown out-of-domains.

\section{Variants of Fourier data augmentation}

The form of Fourier-based data augmentation is not restricted to amplitude swap (AS) or amplitude mix (AM) mentioned in the paper. Here we further propose three more variants of Fourier-based data augmentation:
\begin{itemize}
\item \textbf{Amplitude CutMix (AC)}: We define a pixel-level mixing strategy based on the CutMix~\cite{yun2019cutmix}. Specifically, we firstly sample a binary mask $\mathbf{s}$ from Bernoulli distribution in the shape of the input image. We then linearly mix the amplitude components of two images to generate the augmented images:
\begin{equation}
\hat{\mathcal{A}}(x_{i}^{k}) = (1-\mathbf{s}) \cdot \mathcal{A}(x_{i}^{k}) + \mathbf{s} \cdot \mathcal{A}(x_{i'}^{k'})
\end{equation}
where [$\cdot$] denotes element-wise prodution. Note that AS can be seen as a special case of AC, where the entries of $\mathbf{s}$ in the center area are fixed as 1 and 0 for the remaining area. 

\item \textbf{Amplitude Jittering (AJ)}: We can directly perturb the amplitude information in an image with random noises. Suppose a Gaussian noise $\mathbf{n} \sim \mathcal{N}(0, \sigma)$, we can generate the perturbed amplitude spectrum as:
\begin{equation}
\hat{\mathcal{A}}(x_{i}^{k}) = (1 + \mathbf{n}) \cdot \mathcal{A}(x_{i}^{k})
\end{equation}
where the strength of noise jittering is controlled by $\sigma$. 

\item \textbf{Amplitude Elimination (AE)}: The above Fourier-based data augmentation are all based on amplitude perturbation. Nevertheless, we can directly use the phase-only reconstruction as augmented versions of the original images. In this way, the amplitude information in original images is completely eliminated.
\end{itemize}

We report the performances of all different augmentations incorporated in the baseline DeepAll and our FACT framework in Table~\ref{aug_type_only} and~\ref{aug_type_fact} respectively. Among all the augmentation types, AM performs best in terms of the average performance. AJ with a larger $\sigma$ ($0.5 \sim 0.7$) can also reach a relatively good performance. Interestingly, when incorporated in FACT, the performance of AJ shows a clear trade-off on different domains as the parameter $\sigma$ changes. With a relatively smaller $\sigma$ (0.1 $\sim$ 0.3), AJ performs better when generalizing to the cartoon and photo domain, while with a relatively larger $\sigma$ (0.5 $\sim$ 0.7), AJ is better at generalizing to the art and sketch domain. 
Further increasing the value of $\sigma$ (\eg $\sigma$ = 0.9) will degrade the performance, mainly due to the augmentation strategy is too aggressive. Nevertheless, the AM strategy shows a better trade-off on all the four cases, thus is a more general choice than AJ. 

On the other hand, AC gains a moderate performance among all the perturbation-based variants, which means a simple linear interpolation strategy of AM is a better choice. However, the AE strategy which directly eliminates the amplitude information performs relatively worse than all the other augmentation variants. This may attribute to the large distribution discrepancy of the phase-only reconstructed images compared with the original image domain, which may increase the difficulty of model learning. Another issue is that the phase-only images also follows a different distribution with the pre-trained dataset ImageNet, which means the model may benefit less from ImageNet pre-training. 
Furthermore, as we mentioned in the \textbf{Discussion} of main paper, desirable performances on domains like photo may also require the presence of amplitude information, as these domains contain rich low-level details, therefore totally eliminating amplitude information and overly highlighting phase information may not be a good choice. 

All the above augmentation types are just part of the instantiations of Fourier-based data augmentation. In the future, other more effective instantiations of Fourier-based data augmentation may be proposed. Composition of different augmentation operations is also a topic to be studied. 

\begin{table*}[!htbp]
\centering
\caption{Leave-one-domain-out results on PACS with different variants of Fourier-based data augmentation. The backbone network is ResNet18. The performances are reported from \textit{DeepAll} trained with the augmented images.}
\setlength{\tabcolsep}{6.5mm}{\begin{tabular}{l|cccc|c}
\toprule
Augmentation    & Art        & Cartoon    & Photo      & Sketch     & Avg.  \\ \midrule
AS-partial    & 82.00±0.13 & 76.19±0.15 & 93.89±0.18 & 77.27±1.18 & 82.34 \\
AS-full    & 83.50±0.73 & 76.07±0.30 & 94.49±0.50 & 77.13±2.19 & 82.80 \\ \midrule
AC          & 82.63±0.50 & 77.15±0.48 & 94.94±0.03 & 75.01±0.70 & 82.43 \\ \midrule
AJ ($\sigma$ $=0.1$) & 80.96±0.97 & 76.20±0.59 & 94.55±0.36 & 76.61±0.13 & 82.08 \\
AJ ($\sigma$ $=0.3$) & 81.18±0.22 & 77.71±0.91 & 95.22±0.26 & 79.10±0.04 & 83.30 \\
AJ ($\sigma$ $=0.5$) & 81.62±0.65 & 77.34±0.67 & 94.49±0.40 & 79.24±1.92 & 83.17 \\
AJ ($\sigma$ $=0.7$) & 80.71±0.28 & \textbf{77.72±0.53} & 94.96±0.32 & \textbf{80.22±0.49} & 83.40 \\
AJ ($\sigma$ $=0.9$) & 80.50±0.66 & 76.58±1.07 & 94.01±0.21 & 78.12±1.30 & 82.30 \\\midrule
AE           & 80.08±0.47 & 76.12±0.81 & 93.57±0.45 & 78.50±2.26 & 82.07 \\ \midrule
AM             & \textbf{83.90±0.50} & 76.95±0.45 & \textbf{95.55±0.12} & 77.36±0.71 & \textbf{83.44} \\ \bottomrule
\end{tabular}}
\label{aug_type_only}
\end{table*}

\begin{table*}[!htbp]
\centering
\caption{Leave-one-domain-out results on PACS with different variants of Fourier-based data augmentation. The backbone network is ResNet18. The performances are reported from \textit{FACT} trained with the augmented images.}
\setlength{\tabcolsep}{6.5mm}{\begin{tabular}{l|cccc|c}
\toprule
Augmentation    & Art        & Cartoon    & Photo      & Sketch     & Avg.  \\ \midrule
AS-partial    & 81.61±0.06 & 76.95±0.14 & 93.83±0.61 & 78.30±0.80 & 82.67 \\
AS-full    & 83.46±0.28 & 77.37±0.86 & 94.10±0.34 & 78.63±0.61 & 83.39 \\ \midrule
AC          & 83.32±0.79 & 77.79±0.38 & 94.70±0.14 & 79.24±1.04 & 83.76 \\ \midrule
AJ ($\sigma$ $=0.1$) & 80.66±0.49 & \textbf{78.45±0.99} & \textbf{95.45±0.11} & 77.59±0.60 & 83.04 \\
AJ ($\sigma$ $=0.3$) & 82.62±0.68 & 77.94±0.82 & 95.24±0.27 & 78.74±0.63 & 83.64 \\
AJ ($\sigma$ $=0.5$) & 82.47±0.85 & 77.70±0.64 & 95.12±0.43 & 80.83±0.54 & 84.03 \\
AJ ($\sigma$ $=0.7$) & 83.09±0.12 & 77.22±0.88 & 94.97±0.13 & 80.80±1.16 & 84.02 \\ 
AJ ($\sigma$ $=0.9$) & 82.29±0.30 & 77.23±0.87 & 94.54±0.34 & \textbf{81.16±0.35} & 83.80\\\midrule
AE           & 81.43±0.42 & 76.17±0.22 & 93.78±0.47 & 79.57±0.77 & 82.74 \\ \midrule
AM             & \textbf{85.37±0.29} & 78.38±0.29 & 95.15±0.26 & 79.15±0.69 & \textbf{84.51} \\ \bottomrule
\end{tabular}}
\label{aug_type_fact}
\end{table*}

\section{Sensitivity to different hyperparameters}

In this section we carry out detailed ablation studies about the sensitivity to different hyperparameters related to our method. If not specifically mentioned, all the experiments below are conducted based on the ResNet18 backbone on PACS. When investigating the sensitivity to a specific hyperparameter, other hyperparameters are fixed to their default values, \ie, $m=0.9995$, $T=10$, $(\eta, \beta)=(1.0, 2)$ for PACS and $(\eta, \beta)=(0.2, 200)$ for OfficeHome. 

\subsection{Sensitivity to the momentum $m$}
The results are shown in Table~\ref{sen_momentum}. Basically, a larger momentum value is expected to enhance the effect of the teacher model, thus induce a better performance. Therefore, in all the remaining experiments, we set the momentum value $m$ to 0.9995. 

\begin{table*}[!htbp]
\centering
\caption{Sensitivity to the momentum $m$. Results are reported based on the ResNet18 backbone on PACS.}
\setlength{\tabcolsep}{6.5mm}{\begin{tabular}{l|cccc|c}
\toprule
Momentum & Art                 & Cartoon             & Photo               & Sketch              & Avg.           \\ \midrule
$m=0.9$      & 84.41±0.65          & 76.36±0.84          & 94.64±0.16          & 78.99±1.11          & 83.60          \\
$m=0.99$     & 84.07±1.05          & 77.22±0.88          & 94.61±0.28          & \textbf{79.48±0.98} & 83.84          \\
$m=0.999$    & 84.61±0.06          & 77.84±0.35          & 95.14±0.35          & 79.05±0.71          & 84.16          \\
$m=0.9995$   & \textbf{85.37±0.29} & \textbf{78.38±0.29} & \textbf{95.15±0.26} & 79.15±0.69          & \textbf{84.51} \\ \bottomrule
\end{tabular}}
\label{sen_momentum}
\end{table*}

\subsection{Sensitivity to the temperature $T$}

The results are shown in Table~\ref{sen_temperature}. FACT is not sensitive to the change of the temperature value. Generally, a temperature $T>1$ would lead a good performance. More specifically, a  relatively smaller value of $T$ (\eg $T=1 \sim 5$) would result in a better performance on the sketch domain, but at the sacrifice of the performances on other target domains. While a relatively larger value of $T$ (\eg $T=10$) can reach a better trade-off between all the four leave-one-domain-out cases. Continuously increase the value of $T$ (\eg $T=20$) would degrades the average performance again, mainly because the discriminality in predictions is over-smoothed, thus confusing the model in decision making. A similar trend can also be found on other datasets. Therefore, for convenience, we set the temperature $T=10$ for all the experiments. 

\begin{table*}[!htbp]
\centering
\caption{Sensitivity to temperature $T$. Results are reported based on the ResNet18 backbone on PACS.}
\setlength{\tabcolsep}{6.5mm}{\begin{tabular}{l|cccc|c}
\toprule
Temperature & Art                 & Cartoon             & Photo               & Sketch              & Avg.           \\ \midrule
$T=1$           & 84.06±0.02          & 76.52±0.28          & 93.47±0.12          & \textbf{81.04±0.64} & 83.77          \\
$T=2$           & 84.23±0.52          & 78.11±0.12          & 94.05±0.29          & 80.20±0.92          & 84.15          \\
$T=5$         & 84.46±0.29          & 77.87±0.37          & 95.11±0.25          & 80.49±0.14          & 84.48          \\
$T=10$          & \textbf{85.37±0.29} & \textbf{78.38±0.29} & \textbf{95.15±0.26} & 79.15±0.69          & \textbf{84.51} \\
$T=20$          & 84.91±0.11          & 77.96±0.41          & 94.88±0.21          & 79.16±0.52          & 84.23          \\ \bottomrule
\end{tabular}}
\label{sen_temperature}
\end{table*}

\subsection{Sensitivity to the perturbation strength $\eta$}

\begin{table*}[!htbp]
\centering
\caption{Sensitivity to the perturbation strength $\eta$. Results are reported based on the ResNet18 backbone. For PACS, the consistency loss weight $\beta$ is fixed as 2, and for OfficeHome, $\beta$ is fixed as 200.}
\setlength{\tabcolsep}{6mm}{\begin{tabular}{c|cccc|c}
\toprule
PACS       & Art                 & Cartoon             & Photo               & Sketch              & Avg.           \\ \midrule
$(\eta=0.0, \beta=2)$   & 82.68±0.44          & 78.06±0.39          & 95.35±0.44          & 74.76±0.67          & 82.71          \\
$(\eta=0.2, \beta=2)$   & 82.29±0.51          & 77.73±0.68          & \textbf{96.33±0.25} & 75.10±1.33          & 82.86          \\
$(\eta=0.4, \beta=2)$   & 83.51±0.38          & 77.68±0.46          & 96.23±0.08          & 76.48±0.72          & 83.48          \\
$(\eta=0.6, \beta=2)$   & 83.96±0.51          & \textbf{78.57±0.16} & 95.95±0.11          & 78.06±0.85          & 84.14          \\
$(\eta=0.8, \beta=2)$   & 84.18±0.15          & 78.33±0.41          & 95.72±0.23          & 78.16±0.10          & 84.10          \\
$(\eta=1.0, \beta=2)$   & \textbf{85.37±0.29} & 78.38±0.29          & 95.15±0.26          & \textbf{79.15±0.69} & \textbf{84.51} \\ 
\toprule
OfficeHome & Art                 & Clipart             & Product             & Real                & Avg.           \\ \midrule
$(\eta=0.0, \beta=200)$ & 59.18±0.40          & 54.03±0.61          & 73.91±0.22          & 76.10±0.10          & 65.81          \\
$(\eta=0.2, \beta=200)$ & \textbf{60.34±0.11} & 54.85±0.37          & \textbf{74.48±0.13} & \textbf{76.55±0.10} & \textbf{66.56} \\
$(\eta=0.4, \beta=200)$ & 59.65±0.34          & 55.09±0.21          & 73.87±0.32          & 76.19±0.18          & 66.20          \\
$(\eta=0.6, \beta=200)$ & 58.51±0.38          & \textbf{55.10±0.13} & 73.22±0.23          & 75.48±0.25          & 65.58          \\
$(\eta=0.8, \beta=200)$ & 57.13±0.63          & 55.03±0.58          & 72.93±0.15          & 74.68±0.13          & 64.94          \\
$(\eta=1.0, \beta=200)$ & 57.86±0.14          & 53.25±0.06          & 72.70±0.31          & 74.42±0.26          & 64.56          \\ \bottomrule
\end{tabular}}
\label{sen_eta}
\end{table*}

Recall that the mix coefficient $\lambda$ in the AM strategy are sample from a uniform distribution $U(0, \eta)$, thus the value of $\eta$ controls the strength of the amplitude perturbation. We present the impact of different values of $\eta$ on PACS and OfficeHome in Table~\ref{sen_eta}. Note that when $\eta=0$, no Fourier-based data augmentation is applied, and the difference between the original image and its augmented counterpart\footnote{More exactly, ``the original image and its augmented counterpart'' are two differently augmented versions of the same image.} is only induced by the randomness in basic augmentations (\ie, flipping, cropping, and color jittering).

As we can see, the value of $\eta$ has different effects on different datasets. On PACS, a larger $\eta$ would result in a better performance, and the best performance is achieved at $\eta=1.0$. While on OfficeHome, a smaller $\eta$ does better and the best performance is achieved at $\eta=0.2$. The different behavior of $\eta$ on different datasets can be attribute to the different extent of domain discrepancy. On PACS, the discrepancy between the four domains is much larger than that on OfficeHome. If we treat the Fourier-based augmentation as a kind of regularization, then a larger $\eta$ may induce an overly-regularized model for OfficeHome, considering a vanilla baseline can already perform well due to the small domain discrepancy.  
Therefore, a more aggressive augmentation strategy with a larger $\eta$ will do better on PACS, while a more conservative augmentation strategy with a smaller $\eta$ is more suitable for OfficeHome.

\subsection{Sensitivity to the consistency loss weight $\beta$}

\begin{table*}[!htbp]
\centering
\caption{Sensitivity to the consistency loss weight $\beta$. Results are reported based on the ResNet18 backbone. For PACS, the perturbation strength $\eta$ is fixed as 1.0, and for OfficeHome, $\eta$ is fixed as 0.2.}
\setlength{\tabcolsep}{6mm}{\begin{tabular}{c|cccc|c}
\toprule
PACS       & Art                 & Cartoon             & Photo               & Sketch              & Avg.           \\ \midrule
$(\eta=1.0, \beta=0.0)$ & 83.90±0.50                              & 76.95±0.45                              & \textbf{95.55±0.12}                     & 77.36±0.71                               & 83.44                              \\
$(\eta=1.0, \beta=0.1)$ & 83.70±0.48                              & \textbf{78.54±0.91}                     & 94.85±0.34                              & 77.03±0.95                               & 83.53                              \\
$(\eta=1.0, \beta=1.0)$ & 84.42±0.30                              & 78.30±0.28                              & 95.33±0.30                              & 79.10±0.69                               & 84.29                              \\
$(\eta=1.0, \beta=2.0)$ & \textbf{85.37±0.29} & 78.38±0.29          & 95.15±0.26          & 79.15±0.69          & \textbf{84.51} \\
$(\eta=1.0, \beta=5.0)$ & 84.50±1.00                              & 78.13±0.45                              & 94.81±0.22                              & 79.77±0.33                               & 84.30                              \\
$(\eta=1.0, \beta=10)$  & 84.26±0.07                              & 78.20±0.38                              & 94.77±0.25                              & 80.02±1.23                               & 84.31                              \\
$(\eta=1.0, \beta=20)$  & 84.54±0.49                              & 77.61±0.48                              & 94.62±0.27                              & 80.07±0.72                               & 84.21                              \\
$(\eta=1.0, \beta=200)$ & 82.78±0.47                              & 77.13±0.47                              & 92.81±0.12                              & \textbf{80.23±0.70}                      & 83.24                              \\ 

\toprule
OfficeHome & Art                 & Clipart            & Product             & Real                & Avg.           \\ \midrule
$(\eta=0.2, \beta=0.0)$ & 59.22±0.14                              & 52.97±0.42                              & 73.22±0.06                              & 75.36±0.14                               & 65.19                              \\
$(\eta=0.2, \beta=1.0)$ & 59.38±0.67                              & 53.22±0.23                              & 73.25±0.19                              & 75.28±0.05                               & 65.28                              \\
$(\eta=0.2, \beta=2.0)$ & 59.12±0.28                              & 52.77±0.31                              & 73.66±0.48                              & 75.44±0.30                               & 65.25                              \\
$(\eta=0.2, \beta=5.0)$ & 59.52±0.43                              & 52.97±0.00                              & 73.52±0.14                              & 75.36±0.12                               & 65.34                              \\
$(\eta=0.2, \beta=10)$  & 59.30±0.64                              & 53.13±0.58                              & 73.74±0.39                              & 75.74±0.42                               & 65.48                              \\
$(\eta=0.2, \beta=20)$  & 59.65±0.80                              & 53.46±0.42                              & 73.63±0.22                              & 75.68±0.37                               & 65.60                              \\
$(\eta=0.2, \beta=50)$  & 59.87±0.21                              & 54.10±0.42                              & 74.23±0.19                              & 75.82±0.49                               & 66.00                              \\
$(\eta=0.2, \beta=100)$ & 60.23±0.43                              & 54.15±0.36                              & 74.41±0.25                              & 76.20±0.41                               & 66.25                              \\
$(\eta=0.2, \beta=200)$ & 60.34±0.11          & \textbf{54.85±0.37} & \textbf{74.48±0.13} & \textbf{76.55±0.10} & \textbf{66.56} \\
$(\eta=0.2, \beta=500)$ & \textbf{60.39±0.61}                     & 53.63±0.75                              & 74.18±0.19                              & 76.42±0.24                               & 66.16                              \\ \bottomrule
\end{tabular}}
\label{sen_beta}
\end{table*}

The results of the impact of different values of the consistency loss weight $\beta$ are shown in Table~\ref{sen_beta}. Note that when $\beta=0$, the model is trained without any consistency constraint. The average performance of FACT on PACS is quite stable when $\beta$ is in the range of $1.0 \sim 20$, and the best performance is achieved at $\beta=2.0$. A further smaller or larger value of $\beta$ will either induce a too weak or too strong constraint. On the other hand, FACT performs better with a larger value of $\beta$ on OfficeHome. The performance is stable when $\beta$ is within the range $20 \sim 500$, and the best performance is achieved at $\beta=200$ on OfficeHome. 

An interesting finding is that there seems to be a trade-off between the value of $\eta$ and $\beta$. A larger $\eta$ together with a smaller $\beta$ is better for PACS, while a smaller $\eta$ together with a larger $\beta$ is better for OfficeHome. It seems that $\eta$ and $\beta$ will compensate for each other in terms of regularization power. When equipped with a large $\eta$, using a large $\beta$ may be too aggressive for the model to learn. On the other hand, when equipped with a small $\eta$, still using a small $\beta$ may not fully develop the power of consistency constraint.

In conclusion, for datasets with a large domain discrepancy (\eg PACS, Digits-DG), a larger value of $\eta$ (\eg $\eta=1.0$) together with a smaller value of $\beta$ (\eg $\beta=2.0$) is desired, while for datasets with a small domain discrepancy (\eg OfficeHome), we suggest a smaller value of $\eta$ (\eg $\eta=0.2$) together with a larger value of $\beta$ (\eg $\beta=200$). 

\section{Sampling strategies for amplitude mixing}

\begin{table*}[!htbp]
\centering
\caption{Impact of different sampling strategies for amplitude mixing. Results are reported based on the ResNet18 backbone on PACS.}
\setlength{\tabcolsep}{6.5mm}{\begin{tabular}{c|cccc|c}
\toprule
PACS         &Art &Cartoon & Photo & Sketch & Avg. \\ \midrule
Intra-domain & 83.99±0.10              & \textbf{78.84±0.38}         & \textbf{95.39±0.12}       & 79.01±0.68                  & 84.31                    \\
Inter-domain & 85.09±0.20              & 77.43±0.49                  & 94.71±0.17                & \textbf{79.46±0.92}         & 84.17                    \\
Random       & \textbf{85.37±0.29}     & 78.38±0.29                  & 95.15±0.26                & 79.15±0.69                  & \textbf{84.51}           \\ \bottomrule
\end{tabular}}
\label{from_domain}
\end{table*}

When implementing the amplitude mixing (AM) for a specific image, the amplitude spectrum from another image is sampled randomly from the whole dataset. Nevertheless, we can restrict the sampling image to be taken from the same or different domain. In other words, we can develop an intra-domain AM operation or an inter-domain AM operation. To study the impact of different sampling strategies, we carry out experiments by using only intra-domain or inter-domain AM operations. The results are shown in Table~\ref{from_domain}. 
As we can see, FACT is not sensitive to the sampling strategies. Whether intra-domain or inter-domain sampling strategy brings a good performance, and using a fully random strategy works best, perhaps because more augmented variants are included through fully random sampling.

